\documentclass[10pt,twocolumn,letterpaper]{article}

\usepackage{iccv}
\usepackage{times}
\usepackage{epsfig}
\usepackage{graphicx}
\usepackage{amsmath}
\usepackage{amssymb}
\usepackage{algorithm}
\usepackage{algpseudocode}
\usepackage{makecell}
\usepackage{booktabs}

\usepackage[breaklinks=true,bookmarks=false]{hyperref}

\iccvfinalcopy 

\def\iccvPaperID{12} 

\ificcvfinal\fi
\begin{document}

\title{Visual Tracking by means of Deep Reinforcement Learning \\ and an Expert Demonstrator}

\author{Matteo Dunnhofer, Niki Martinel, Gian Luca Foresti and Christian Micheloni\\
Department of Mathematics, Computer Science and Physics, 
University of Udine, Italy\\
{\tt\footnotesize dunnhofer.matteo@spes.uniud.it},
{\tt\footnotesize \{niki.martinel, gianluca.foresti, christian.micheloni\}@uniud.it}
}

\maketitle
\thispagestyle{empty}

\begin{abstract}
   In the last decade many different algorithms have been proposed to track a generic object in videos. Their execution on recent large-scale video datasets can produce a great amount of various tracking behaviours. New trends in Reinforcement Learning showed that demonstrations of an expert agent can be efficiently used to speed-up the process of policy learning. Taking inspiration from such works and from the recent applications of Reinforcement Learning to visual tracking, we propose two novel trackers, A3CT, which exploits demonstrations of a state-of-the-art tracker to learn an effective tracking policy, and A3CTD, that takes advantage of the same expert tracker to correct its behaviour during tracking.  Through an extensive experimental validation on the GOT-10k, OTB-100, LaSOT, UAV123 and VOT benchmarks, we show that the proposed trackers achieve state-of-the-art performance while running in real-time.
\end{abstract}

\section{Introduction}
Visual object tracking is one of the most challenging problems in Computer Vision. In its simplest form, it consists in the persistent recognition and localization --by means of bounding boxes-- of a target object in consecutive video frames.
Even though many efforts have been recently made (\eg \cite{Smeulders2014,OTB,VOT2018}), the process of automatically following a generic object in a video comes with several different challenges including occlusions, light changes, fast motion, and motion blur.
In addition, many practical applications of visual tracking, such as video surveillance, behavior understanding, autonomous driving and robotics, require accurate predictions with real-time constraints.

In many current methodologies (\eg ~\cite{Nam2015,Held2016,Gordon2018,Chen2018}), Convolutional Neural Networks (CNNs) \cite{Lecun98} pre-trained for image classification showed to be effective for visual tracking. Due to their discriminative power, CNN generated feature representations are widely used to search the target in the consecutive frames. This kind of information is widely used in classification or tracking-by-detection methods (\eg \cite{Nam2015,Song2017}).
The most significant drawback of these methods is that they require computational demanding procedures to search for candidate targets in new frames. Furthermore, even strong CNN models may not be able to capture all possible variations of targets and need to be updated online during tracking. In these scenarios, the tracker shall understand the quality of its tracking process and the target's motion status, in order to take decisions to efficiently update its model. Solutions that implement such a mechanism achieve excellent results (\eg ~\cite{Nam2015,Danelljan2017,Yun2017,Ren2018}), but their processing speed is often far from being real-time. Moreover, the problem of taking decisions online requires algorithms capable of deciding intelligently at the right moments.

To address these issues, tracking methodologies based on Reinforcement Learning (RL) have been recently proposed~\cite{Choi2017,Yun2017,Supancic2017,Chen2018,Ren2018}. The idea behind such works is to treat aspects like target searching procedures or tracking status evaluation as sequential decision-making problems. In these settings, an artificial agent is trained to take optimal sequential decisions to solve a tracking related task which, ultimately, leads to the development of a strategy to track the target object. 
These solutions maintain competitive performance with state-of-the-art methods, however they implement complex and demanding online update procedures that slow tracking. In addition, these methods are usually not end-to-end and require at least two training stages, one initial supervised learning (SL) stage and a following RL fine-tuning. 

We argue that better speed performance can be obtained and that SL can be incorporated into an RL framework to make the training end-to-end. 
We claim that tracking demonstrations of an expert tracker can be used to guide RL tracking agents. Furthermore, we propose to simplify the online update strategy by taking advantage of the expert during tracking. We will demonstrate that RL functions needed for training can be also used during tracking to exploit the performance of the expert tracker and to consequentially improve the tracking accuracy.

In particular, in this paper we introduce the following contributions:
\begin{enumerate}
    \item a real-time CNN-based tracker named A3CT which is trained via an end-to-end RL method that takes advantage of the demonstrations of a state-of-the-art tracker;
    \item a real-time CNN-based tracker named A3CTD which uses the RL functions learned during training to improve performance by exploiting the expert during the tracking phase.
\end{enumerate}

The proposed trackers are built on a deep regression network for tracking \cite{Held2016,Gordon2018} and are trained inside an on-policy Asynchronous Actor-Critic framework \cite{Mnih2016} that incorporates SL and expert demonstrations. A state-of-the-art tracking algorithm \cite{Bertinetto2016} is run on a large-scale tracking dataset \cite{GOT10k} to obtain the demonstrations. Experiments will show that the proposed A3CT and A3CTD trackers perform comparably with state-of-the-art methods on the GOT-10k test set \cite{GOT10k}, LaSOT \cite{LaSOT}, UAV123 \cite{UAV123}, OTB-100 \cite{OTB} and VOT benchmarks \cite{VOT2016,VOT2018}, while achieving a processing speed of 90 FPS and 50 FPS respectively.

\section{Related work}

\subsection{Deep RL}
RL concerns methodologies to train artificial agents to solve interactive decision-making problems \cite{SuttonBarto2018}.  
Recent trends in this field (\eg~\cite{Mnih2013,Mnih2015,Silver2016,Silver2017}) showed the successful combination of Deep Neural Networks (DNNs) and RL algorithms (so-called Deep RL) in the representation of models such as the value or policy functions. Among the existing approaches, off-policy strategies aim to learn the state or the state-action value functions, that give estimations about the expected future reward of states and actions \cite{Watkins1992,Mnih2013,Mnih2015}. The policy is then extracted by choosing greedily the actions that yield the highest function values.
On the other hand, on-policy algorithms directly learn the policy by optimizing the DNN with respect to the expected future reward \cite{Williams1992}.
There exist then hybrid approaches, known as Actor-Critic \cite{Konda2000}, that maintain and optimize the model representations of both the policy and state value (or state-action value) functions.

All these methods however suffer of slow convergence, especially in cases where continuous or high-dimensional action spaces are considered. Recent solutions (\eg \cite{Vecerik2017,Nair2018,Salimans2018,Kang2018,Hester2018}) propose to use expert demonstrations to help and guide the learning process.

\subsection{Visual Tracking}
Visual Tracking has received increasing interest thanks to the introduction of new benchmarks~\cite{OTB,LaSOT,UAV123,GOT10k} and challenges \cite{VOT2014,VOT2015,VOT2016,VOT2017,VOT2018}.
Thanks to their superior representation power, in recent years various approaches based on CNNs appeared \cite{Held2016,Gordon2018,Nam2015,Danelljan2017,Bertinetto2016,Li2018b,Zhang2019}. 
Held \etal \cite{Held2016} and Gordon \etal \cite{Gordon2018} showed how deep regression CNNs could capture the target's motion. 
However, these methods are trained using SL which optimizes parameters for just local predictions. In contrast, we propose a RL-based training which optimizes the DNN's weights for the maximization of performance in future predictions.
Nam \etal \cite{Nam2015} proposed an online tracking-by-detection approach 
by using a pre-trained CNN for image classification.  Similarly, Danelljan \etal \cite{Danelljan2016,Danelljan2017} proposed a discriminative correlation filter approach by integrating multi-resolution CNN features. These solutions obtained outstanding results w.r.t. the previous methodologies, however they are very computationally expensive and can run at just 1 and 6 FPS respectively.
Currently, the approach based on the Siamese framework is getting significant attention for their well-balanced tracking accuracy and efficiency \cite{Bertinetto2016,Guo2017,Li2018,Li2018b,Zhu2018,Zhang2019}. These trackers formulate the visual tracking as a cross-correlation problem and are leveraging effectively from end-to-end learning of DNNs. However their performance is susceptible to visual distractors due to the non-incorporation of temporal information or online fine-tuning. Conversely to this, our tracker present the use of an LSTM \cite{Hochreiter1997} to model the temporal relation of target's appearance between frames.

\subsection{Deep RL for Visual Tracking}
Very recently, Deep RL has started to be increasingly used to tackle the Visual Tracking problem. 
The first solution in this direction was the work of Yun \etal \cite{Yun2017}, which proposed an Action-Decision network to learn a policy for selecting a discrete number of actions to modify iteratively the bounding box in the previous frame.
Huang \etal \cite{Huang2017} used a Deep-Q-Network \cite{Mnih2015} to learn a policy for adaptively selecting efficient image features during the tracking process. 
In the work of \cite{Supancic2017}, the tracker was modeled as an agent that takes decisions during tracking whether: to continue tracking with a state-of-the-art tracker or to re-initialize it; and to update or not the appearance model of the target object. 
In \cite{Choi2017}, authors used a variant of REINFORCE \cite{Williams1992} to develop a template selection strategy, encouraging the tracking agent to choose, at every frame, the best template from a finite pool of candidate templates.
In \cite{Ren2018}, authors presented a tracker which, at every time step, decides to shift the current bounding box while remaining on the same frame, to stop the shift process and move to the next frame, to update on-line the weights of the model or to re-initialize the tracker if the target is considered lost. 
Finally, \cite{Chen2018} proposed to substitute the discrete action framework of \cite{Yun2017} with continuous actions, thus performing just a single action at every frame. 

All the presented methods include a pre-training step that uses SL to build a baseline policy or some other module used later by the tracking agents. Only after, RL is used to fine-tune such policies and modules. We take inspiration from RL methods that exploit expert demonstrations and we propose a novel end-to-end methodology based on on-policy Actor-Critic framework \cite{Mnih2016} to train a DNN capable of tracking generic objects in videos. We also demonstrate that the state value function learned during training, can be directly used to exploit the expert during tracking, in order to adjust wrong tracking behaviors and to consequentially improve the tracking accuracy.

\section{Methodology}
The key idea of this paper is to take advantage of an expert tracker for training and tracking. RL and expert demonstrations are used to train a DNN which is then capable of tracking autonomously a generic target object in a video. The same network is also capable of evaluating its own performance and the one of the expert, thus exploiting the latter's knowledge in potential failure cases. 


\begin{figure*}[!h]
\begin{center}
\includegraphics[width=\linewidth]{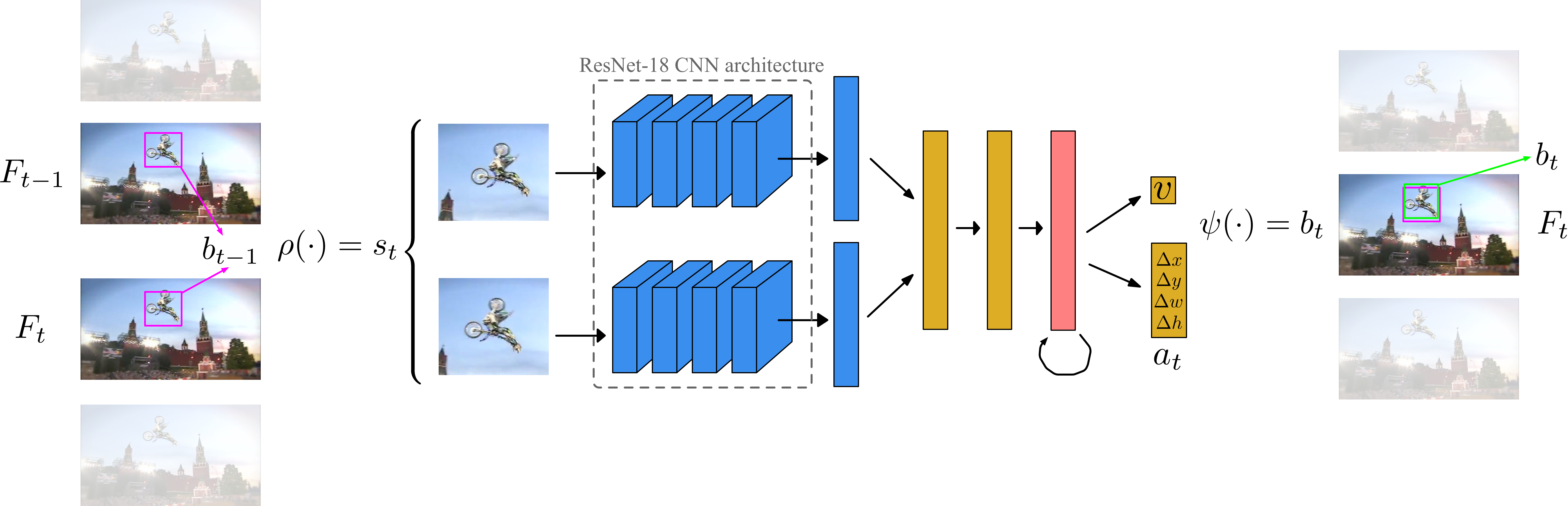}
\end{center}
   \caption{Visual representation of the interaction between the tracking agent and a video. Each pair of frames $F_{t-1}, F_t$ is cropped by the function $\rho(\cdot)$ using the bounding box $b_{t-1}$. The obtained state $s_t$ is fed to the agent's DNN which is composed by two branches of convolutional layers (the blue boxes) followed by, two fully-connected layers (rectangles in yellow), an LSTM layer (in light red) and two other fully connected layers for the prediction of $v$ and the action $a_t$. Finally, the output bounding box $b_t$ is built by the function $\psi(\cdot)$ which moves $b_{t-1}$ by the relative shift $a_t$.}
\label{fig:architecture}
\end{figure*}

\subsection{Problem setting}
\label{sec:votmdp}
In our setting, the tracking problem follows the definition of a Markov Decision Process (MDP). The tracker is treated as an artificial agent which interacts with an environment that is obtained as an MDP defined over a video. MDPs are a standard formulation for RL tasks and are composed of: a set of states $\mathcal{S}$; a set of actions $\mathcal{A}$; a state transition function $f : \mathcal{S} \times \mathcal{A} \rightarrow \mathcal{S}$; a reward function $r : \mathcal{S} \times \mathcal{A} \rightarrow \mathbb{R}$; and a discount value $\gamma \in \mathbb{R}$.

The interaction with the video, which we call an episode,  happens through a temporal sequence of observations $s_1, s_2, \cdots, s_t$, actions $a_1, a_2, \cdots, a_t$ and rewards $r_1, r_2, \cdots, r_t$. In the $t$-th frame, the agent is provided with the state $s_t$ and outputs the continuous action $a_t$ which consists in the relative motion of the target object, i.e. it indicates how its bounding box, which is known in frame $t-1$, should move to enclose the target in the frame $t$. This approach is similar to the MDP formulation given by Chen et al. \cite{Chen2018}, however we propose different definitions for the states, actions and rewards.

\paragraph{Preliminaries.} Given a dataset $\mathcal{D} = \{\mathcal{V}_0, \cdots,\mathcal{V}_{|\mathcal{D}|}\}$, we consider the $j$-th video
\begin{equation}
 \mathcal{V}_j = \big\{ F_t \in \{0,\cdots,255\}^{w \times h \times 3} \big\}_{t=0}^{T_j}
\end{equation}
as a sequence of frames $F_t$. 
 Let $b_t = [x_t,y_t,w_t,h_t]$ be the $t$-th bounding box defining the coordinates of the top left corner, and the width and height of the rectangle that contains the target object.
At time $t-1$, given $F_{t-1}$ and $b_{t-1}$, the goal of the tracker is to predict the bounding box $b_{t}$ that best fits the target in the consecutive frame $F_{t}$.

\paragraph{State.} Every state $s_t \in \mathcal{S}$ is defined as a pair of image patches obtained by cropping frames $F_{t-1}$ and $F_t$ using the bounding box $b_{t-1}$.
Specifically, $s_t = \rho(F_{t-1}, F_t, b_{t-1}, k)$, where $\rho(\cdot)$ crops the frames $F_{t-1}, F_t$ within the area of the bounding box $b_{t-1}' = [x'_{t-1},y'_{t-1},k \cdot w_{t-1},k \cdot h_{t-1}]$ that has the same center coordinates of $b_{t-1}$ but which width and height are scaled by $k$. With this function and by choosing $k > 1$, we can control the amount of additional image context information that is provided to the agent. 

\paragraph{Actions and State Transition.} Each action $a_t \in \mathcal{A}$ consists in a vector $a_t = [\Delta x_t, \Delta y_t, \Delta w_t, \Delta h_t] \in [-1,1]^4$ which defines the relative horizontal and vertical translations ($\Delta x_t, \Delta y_t$, respectively) and width and height scale variations ($\Delta w_t, \Delta h_t$, respectively) that have to be applied to $b_{t-1}$ to predict the bounding box $b_t$. This is obtained through $\psi: \mathcal{A} \times \mathbb{R}^4 \rightarrow \mathbb{R}^4$ such that
\begin{align}
    \psi(a_t, b_{t-1}) = 
    \begin{cases}
    x_t = x_{t-1} + \Delta x_t \cdot w_{t-1}\\
    y_t = y_{t-1} + \Delta y_t \cdot h_{t-1}\\
    w_t = w_{t-1} + \Delta w_t \cdot w_{t-1}\\
    h_t = h_{t-1} + \Delta h_t \cdot h_{t-1}\\
    \end{cases}
\end{align}
After performing the action $a_t$, the agent moves from the state $s_t$ into the state $s_{t+1}$ which is defined as the pair of cropped images obtained from the frames $F_t$ and $F_{t+1}$ using the bounding box $b_t$.

\paragraph{Reward.} The reward function $r(s_t, a_t)$ expresses the quality of the action $a_t$ taken at state $s_t$. 
Our reward definition is based on the Intersection-over-Union (IoU) metric computed between $b_t$ and the ground-truth bounding box, denoted as $g_t$, \ie, $\text{IoU}(b_t, g_t) = (b_t \cap g_t) / (b_t \cup g_t) \in [0,1]$. At every interaction step $t$, the reward is formally defined as
\begin{align}
r(s_t, a_t) = 
    \begin{cases}
    \omega\left(\text{IoU}(b_t, g_t)\right) \ \text{if IoU}(b_t, g_t) \geq 0.5 \\
    -1 \ \text{otherwise}
    \end{cases}
\end{align}
with
\begin{equation}
\omega(z) = 2(\lfloor z \rfloor_{0.05}) - 1
\end{equation}
flooring to the closest $0.05$ digit, then shifting the input range from $[0,1]$ to $[-1,1]$. 

\paragraph{Expert demonstrations.} To guide the learning of our tracking agent we take advantage of the positive demonstrations of an expert tracker.
Given $\mathcal{V}_j$, the bounding box prediction of the expert at time $t$ is denoted as $b^{(d)}_t$.
The demonstrations are obtained as sequences of triplets $\{(s^{(d)}_t, a^{(d)}_t, r^{(d)}_t)\}_{t=0}^{T_j}$, each containing a state, an action and a reward, respectively.
Precisely, we have that $s^{(d)}_t = \rho(F_{t-1},F_{t},b^{(d)}_{t-1},k)$ and $a^{(d)}_t = [\Delta x^{(d)}_t, \Delta y^{(d)}_t, \Delta w^{(d)}_t, \Delta h^{(d)}_t]$, where its elements are obtained through $\phi : \mathbb{R}^4 \times \mathbb{R}^4 \xrightarrow{} \mathcal{A}$, defined as  
\begin{align}
\phi(b^{(d)}_{t}, b^{(d)}_{t-1}) = 
    \begin{cases}
    \Delta x^{(d)}_t = (x^{(d)}_{t} - x^{(d)}_{t-1}) / w^{(d)}_{t-1} \\
    \Delta y^{(d)}_t = (y^{(d)}_{t} - y^{(d)}_{t-1}) / h^{(d)}_{t-1} \\
    \Delta w^{(d)}_t = (w^{(d)}_{t} - w^{(d)}_{t-1}) / w^{(d)}_{t-1} \\
    \Delta h^{(d)}_t = (h^{(d)}_{t} - h^{(d)}_{t-1}) / h^{(d)}_{t-1} \\
    \end{cases}
\end{align}
Rewards are calculated as $r^{(d)}_t = r(s^{(d)}_t, a^{(d)}_t)$.



\subsection{Agent architecture}
Our tracking agent maintains representations of both the policy $\pi : \mathcal{S} \rightarrow \mathcal{A}$ and the state value function $v : \mathcal{S} \rightarrow \mathbb{R}$. This is done by using a DNN with parameters $\theta$. In particular, we used a deep architecture that is similar to the one proposed by Gordon \etal \cite{Gordon2018}. 

The network gets as input two image patches. These pass through two convolutional branches that have the form of ResNet-18 CNN architecture \cite{He2016} and which weights are pre-trained for image classification on the ImageNet dataset \cite{ImageNet}. The two tensors of feature maps produced by the branches are first linearized, then concatenated together and finally fed to two consecutive fully connected layers with ReLU activations. After that, the features are inputted to an LSTM \cite{Hochreiter1997} RNN. Both the fully connected layers and the LSTM are composed of 512 neurons. The output of the LSTM is finally fed to two separate fully connected heads, one that outputs the action $a_t = \pi(s_t | \theta)$ and the other that outputs the value of the state, i.e. $v(s_t | \theta)$.

In Figure \ref{fig:architecture} a visual representation of the DNN architecture, together with the interaction process, is presented.

\subsection{Training}
The proposed DNN is trained solely off-line and in an end-to-end manner. The implemented training procedure is based on the on-policy A3C \cite{Mnih2016} RL framework. This method exploits $P$ parallel and independent agents that interact with their own environments and that later use the gained experience to update asynchronously the weights $\theta$ which are shared among all agents. Indeed, each agent owns a copy $\theta'$ of the weights and this is synchronized with $\theta$ after every learning step. A3C is a standard algorithm in RL, however it is not designed to take expert demonstrations into account. To overcome this limitation for our problem, we set up an A3C framework where a first half of the learning agents performs the traditional A3C learning, while the other half learns to imitate the actions of the expert tracker demonstrator in a supervised fashion. 

\paragraph{Imitating agents.} Each imitating agent interacts with its environment by observing states, performing actions and receiving rewards just as standard A3C agents. Every $t_{max}$ steps the agent updates the weights $\theta$ of the shared model with the gradients of the following loss function
\begin{align}
\label{eq:l1loss}
\mathcal{L}_{imit} = \sum_{i=1}^{t_{max}} |\phi(b^{(d)}_t, b_{t-1}) - a_t| \cdot m_i.
\end{align}
which is the L1 loss between the actions performed by the learning agent and the actions that the expert tracker would take to move the agent's bounding box $b_{t-1}$ into the expert's $b^{(d)}_t$. These absolute values are masked by the values $m_i \in \{0,1\}$. Each of these is computed during the interaction and determines the situation in which the agent performed worse than demonstrator ($m_i = 1$) or better ($m_i = 0$). By optimizing the loss function \ref{eq:l1loss}, the weights $\theta$ are changed only if the agent's performance, in terms of received reward, is lower than the performance of the expert tracker. In simple words, the demonstrator is used to learn a baseline behavior on which the RL agent can build up its own tracking strategy, thus reducing the random exploration and consequentially speed up the learning process. 

\paragraph{RL agents.} The training process performed by RL agents follows the standard structure proposed by Mnih et al. \cite{Mnih2016} for continuous control. Each agent interacts with the environment for a maximum of $t_{max}$ steps. However, differently from the imitating agents, at each step $t$ the RL agents sample actions from a normal distribution $\mathcal{N}(\mu, \sigma)$, where the mean is the predicted action, $\mu = \pi(s_t | \theta')$, and the standard deviation is obtained as $\sigma = |\pi(s_t) - \phi(g_t, b_{t-1}) |$ (which is the absolute value of the difference between the agent's action and the action that obtains, by shifting $b_{t-1}$, the ground-truth bounding box $g_t$). Intuitively, $\sigma$ shrinks $\mathcal{N}$ when the action $a_t$ is close to the ground-truth action $\phi(g_t, b_{t-1})$, thus reducing the chance of choosing potential wrong actions when approaching the correct one. On the other hand, when the action $a_t$ is far from $\phi(g_t, b_{t-1})$, $\sigma$ takes a greater value, spreading $\mathcal{N}$. This allows the agent to explore more the environment and discover potential good actions. 

\paragraph{Curriculum strategy.} In addition to the guiding process done by the imitating learners using the expert demonstrations, we designed a curriculum learning strategy \cite{Bengio2009} to further facilitate the training. In a similar way as proposed by \cite{Salimans2018}, we built a curriculum based on the performance of the learning agents w.r.t. to the expert demonstrator. In particular, after terminating each episode, a success counter is increased if the agent performs better than the expert in that episode, i.e. if the former's cumulative reward, received up to $\widehat{T}_j$, is greater or equal to the one obtained by the latter. In formal terms, the success counter is updated if the following holds
\begin{align}
\sum_{i=1}^{\widehat{T}_j} r_i \geq \sum_{i=1}^{\widehat{T}_j} r^{(d)}_i.
\end{align}
The counter update is done by testing agents that interact with the sequences by performing $\pi(s_t | \theta')$ using a local copy $\theta'$ of the shared weights.
The terminal episode index $\widehat{T}_j$ is successively increased during the training procedure by a central process which checks if the ratio, between the number of episodes in which the learning agent performs better than the demonstrator and the total number of episodes terminated, is above the threshold $\tau$. 
With this learning setting, we ensure that at every augmentation of $\widehat{T}_j$ the agents face a simpler learning problem where they are likely to succeed and in a shorter time, since they have already developed a tracking policy that, up to $\widehat{T}_j -1$, is at least good as the one of the expert.

\subsection{Tracking at test time}
Despite the fact that the proposed tracker is trained by taking advantage of an expert's knowledge, our tracker develops a tracking ability that can be exploited independently from the tracking strategy used by the demonstrator. Nevertheless, it is possible to take advantage of the expert's tracking performance also during the tracking phase. 
Therefore we set up two tracking strategies, the first one that tries to track autonomously the target object and we refer it as A3CT, and the second one that takes advantage of the demonstrator's knowledge also during tracking and that we name A3CTD. 

\paragraph{A3CT.}
In this setting, A3CT is applied straight away on an arbitrary sequence. Each tracking sequence $\mathcal{V}_j$, with target object outlined by $g_0$, is considered as the MDP described in section \ref{sec:votmdp}. The tracker computes states $s_t$ from frames $F_t$, performs actions as by means of the learned policy $a_t = \pi(s_t | \theta)$ which are used to output the bounding boxes $b_t = \phi(a_t, b_{t-1})$. At the beginning, $b_0 := g_0$.

No online update of the network's weights nor of the LSTM's hidden state are performed. 

\paragraph{A3CTD.}
During training, the tracking agent learns both the policy $\pi(s_t | \theta)$ and the value function $v(s_t | \theta)$. 
$v(\cdot)$ is a function that predicts the reward that the agent expects to receive from the current state $s_t$ to the end of the sequence. Since our reward definition is a direct measure of the IoU between the predictions of the agent and the ground-truth bounding boxes, $v(s_t | \theta)$ gives an estimate of the total amount of IoU that the tracker expects to obtain from state $s_t$ on wards. This function can be exploited as a performance evaluation for both our tracker and the expert demonstrator. 
In particular, at each time step $t$, $\widehat{R} = v(s_t | \theta)$ and $\widehat{R}^{(d)} = v(s^{(d)}_t | \theta)$ are obtained as the evaluation for A3CTD and the expert tracker respectively. The expert state $s^{(d)}_t$ is obtained by cropping frames $F_{t-1}, F_t$ using its previous prediction $b^{(d)}_{t-1}$. By comparing $\widehat{R}$ and $\widehat{R}^{(d)}$, our strategy decides if to output the bounding box of A3CTD or the bounding box produced by the expert tracker. More formally, if $\widehat{R} \geq \widehat{R}^{(d)}$ then the tracker outputs $b_t := \phi(a_t, b_{t-1})$ otherwise it outputs $b_t := b^{(d)}_{t}$.


\subsection{Implementation details}
\label{sec:impldet}
In this section we report the results of the hyperparameters search which led to the best performance.

Before being fed to the DNN, the image crops that forms the MDP states are resized to $[128 \times 128 \times 3]$ pixels and standardized, per channel, by subtracting the mean and dividing by the standard deviation calculated on the ImageNet dataset \cite{ImageNet}. The dilating factor $k$ is set to $1.5$.

A total number of $P = 16$ training agents was used. The discount factor $\gamma$ was set to 1. The length of the rollout was defined in $t_{max}= 5$ steps. $\tau$ was set to 0.25.
The model was trained for $40 000$ episodes using the Adam optimizer~\cite{Kingma2014}. The learning rate for both imitating and training agents was set to $10^{-6}$. A weight decay of $10^{-4}$ was also added to the L1 loss of the imitating agents as regulatory term. 

Training and experiments have been conducted running our Python code with the PyTorch~\cite{Paszke2017} machine learning library on an Intel Xeon E5-2690 v4 @ 2.60GHz CPU with 320 GB of RAM, four NVIDIA TITAN V GPUs and an NVIDIA TITAN Xp GPU each with 12 GB of memory. The training took around 4 days. In the evaluation of trackers' speed, we ignore disk read times since they do not dependent on the tracking algorithm.

\paragraph{Expert Tracker.} The role of expert tracker was assigned to SiamFC \cite{Bertinetto2016}. The choice was motivated by the fact that this solution is nowadays an established methodology in the visual tracking panorama, and it shows great balance in results across many different benchmarks. In particular, SiamFC has currently one of the best performance on the public leader-board of the GOT-10k test set. Additionally, the source code was publicly available. 

To obtain tracking demonstrations, we ran SiamFC on the training set of GOT-10k dataset \cite{GOT10k}. The implemented SiamFC was trained on the ImageNet VID dataset \cite{ImageNet}. This is an important aspect because, to train our tracking agent, we want examples of the tracker's real behaviour, that must be obtained on never seen before sequences. Moreover, demonstrations that are clearly useful are needed. So, of all the trajectories produced, we retained just the ones considered positive, i.e. the trajectories that satisfy $\text{IoU}(b^{(d)}_t, g_t) > 0.5$ for all $t \in \{1,\dots,T_j\}$. All the others were discarded.

\paragraph{Training Dataset.} To train A3CT and A3CTD we leveraged of the training set of the GOT-10k dataset \cite{GOT10k}. This is a large-scale dataset containing 9335 training videos, 180 validation videos and other 180 videos for testing. In total, this dataset provides 1.5M bounding boxes that identify 10k different target objects. The latters belong to 563 distinct object classes. 
The actual number of training sequences we used is however inferior. In fact, just the videos which obtained a positive demonstration from the expert tracker were employed for training. Furthermore, as we aimed to take part to the VOT 2019 challenge, we removed 1000 sequences from the training set. These overlapped with the pool of videos used by the VOT committee for evaluation. After these pruning steps, the total amount of training samples, $|\mathcal{D}|$, resulted in 1782 videos.

\section{Experiments}
In this section we report the experimental setup and we discuss the results, obtained by the proposed trackers A3CT and A3CTD, on the benchmarks GOT-10k \cite{GOT10k}, LaSOT \cite{LaSOT}, UAV123 \cite{UAV123}, OTB-100 \cite{OTB}, VOT-2018 \cite{VOT2018} and VOT-2019.

\subsection{GOT-10k Test Set}
The GOT-10k \cite{GOT10k} test set comprises 180 videos. Target objects belong to 84 different classes and 32 forms of object motion are present. 
To ensure a fair evaluation, the trackers that are evaluated on this benchmark are forbidden from using external datasets for training. The evaluation protocol proposed by the authors is the one-pass evaluation (OPE) \cite{OTB}. The metrics used are the average overlap (AO) and the success rates (SR) with overlap thresholds $0.5$ and $0.75$. 

\begin{table}[!h]
	\centering\vspace{-1mm}
	\resizebox{1.01\columnwidth}{!}{%
		\begin{tabular}{l@{~}c@{~~}c@{~~}c@{~~}c@{~~}c@{~~}c@{~~}c@{~~}c@{~~}c@{~~}c@{~~}}
    
	\toprule
	
	& KCF & MDNet & ECO & CCOT & GOTURN & SiamFC & SiamFCv2 & ATOM & \textbf{A3CT} & \textbf{A3CTD} \\
	& \cite{Henriques2014} & \cite{Nam2015} & \cite{Danelljan2017} & \cite{Danelljan2016} & \cite{Held2016} & \cite{Bertinetto2016} & \cite{Valmadre2017} & \cite{Danelljan2019} &&\\
	
	\midrule
	
	AO & 0.203 & 0.299 & 0.316 & 0.325 & 0.347 & 0.348 & 0.374 & 0.556 &   \textbf{0.415} & \textbf{0.425} \\
	SR$_{0.50}$ & 0.177 & 0.303 & 0.309 & 0.328 & 0.375 & 0.353 & 0.404 & 0.634 & \textbf{0.477} & \textbf{0.495} \\
	SR$_{0.75}$ & 0.065 & 0.099 & 0.111 & 0.107 & 0.124 & 0.098 & 0.144 & 0.402 & \textbf{0.212} & \textbf{0.205} \\

    \bottomrule
    
    \end{tabular}

	}\vspace{1mm}%
\scriptsize

    \caption{State-of-the-art comparison on the GOT-10k test set in terms of average overlap (AO), and success rates (SR) with overlap thresholds $0.5$ and $0.75$. Except for ATOM, both versions of our approach outperform the previous methods in all three measures.}
    \label{tab:got10ksota}%
	\vspace{-1mm}
\end{table}

In Table \ref{tab:got10ksota} we report the results of A3CT and A3CTD against the state-of-the-art. 
A3CT outperforms the state-of-the-art trackers which, at the time of writing, appear on the GOT-10k test set leaderboard. In particular, it has a better tracking performance w.r.t. to the demonstrator tracker SiamFC \cite{Bertinetto2016}, with a performance gain of 6.7\% and in AO, 12.4\% in SR$_{0.50}$, and 11.4\% in SR$_{0.75}$. A3CTD increases additionally the performance of A3CT, with an improvement of 1\% in AO, 1.8 in SR$_{0.50}$ but with a loss of 0.7\% in SR$_{0.75}$.
We perform worse than ATOM \cite{Danelljan2019}, however we remark that these results are obtained considering just 1782 of the 9335 sequences (19\%) contained in the GOT-10k training set. 



\subsection{OTB-100}
\label{sec:otb100res}
The OTB-100 \cite{OTB} benchmark is a set of 100 challenging videos and it is widely used in the tracking literature. The standard evaluation procedure for this dataset is the OPE method and the metrics used are the success plot 
and the precision plot. 
 The Area Under the Curve (AUC) of these curves are referred as success score (SS) and precision scores (PS) respectively.

\begin{table}[!h]
	\centering\vspace{-1mm}
	\resizebox{1.01\columnwidth}{!}{%
		\begin{tabular}{l@{~}c@{~~}c@{~~}c@{~~}c@{~~}c@{~~}c@{~~}c@{~~}c@{~~}c@{~~}c@{~~}}
    
	\toprule
	
	& GOTURN & RE3 & KCF & SiamFC & ACT & MDNet & ECO & SiamRPN++ &\textbf{A3CT} & \textbf{A3CTD} \\
	& \cite{Held2016} & \cite{Gordon2018} & \cite{Henriques2014} & \cite{Bertinetto2016} & \cite{Chen2018} & \cite{Nam2015} & \cite{Danelljan2017} & \cite{Li2018b}&\\
	
	\midrule
	
	SS \quad & 0.395 & 0.464 &0.477 & 0.575 & 0.625 & 0.677 & 0.691 & 0.696 &  \textbf{0.419} & \textbf{0.535} \\
	PS \quad & 0.534 & 0.582 & 0.693 & 0.762 & 0.859 & 0.909 & 0.910 & 0.914  &\textbf{0.568} & \textbf{0.717} \\
	
	\bottomrule
	
    \end{tabular}
	}\vspace{1mm}%
\scriptsize

    \caption{State-of-the-art comparison on the OTB-100 benchmark in terms of success score (SS) and precision score (PS).}
    \label{tab:otb100sota}%
	\vspace{-1mm}
\end{table}

In Table \ref{tab:otb100sota} we report the success and and precision scores against state-of-the-art solutions. On this benchmark, A3CT and A3CTD have lower performance than ECO \cite{Danelljan2017}, MDNet \cite{Nam2015}, SiamRPN++ \cite{Li2018b} and the expert SiamFC \cite{Bertinetto2016}. However, A3CT still performs better than GOTURN \cite{Held2016}. A3CTD instead outperforms RE3 \cite{Gordon2018} and KCF \cite{Henriques2014}, with a  5.8-7.1\% performance gain in SS and 1.8-13.5\% in PS. In this setting, the help of the expert tracker is crucial to improve the results of A3CT, which sees an improvement of 11.6\% in SS and 14.9\% in PS.


\subsection{LaSOT}
We performed evaluations of A3CT and A3CTD performance on the test set of LaSOT benchmark \cite{LaSOT}. This dataset is composed of 280 videos with a total of more than 650k frames and an average sequence length of 2500 frames. 
 To evaluate our tracker, we use the same methodology and metrics used for the OTB-100 experiments. 

In Table \ref{tab:lasotsota} we present the results against state-of-the-art trackers. In this setting, in terms of SS A3CT performs comparably to ECO \cite{Danelljan2017} and RE3 \cite{Gordon2018} but much better than GOTURN \cite{Held2016}. Also in this case, the aid of the expert tracker is crucial, which results in a increment of 10.9\% in SS and of 12.2\% in PS. A3CTD so outperforms the expert SiamFC \cite{Bertinetto2016} in SS by 7.9\% and MDNet \cite{Nam2015} by 1.8\%. 
Both our trackers are however weaker than SiamRPN++ \cite{Li2018b}.

\begin{table}[!h]
	\centering\vspace{-1mm}
	\resizebox{1.01\columnwidth}{!}{%
		\begin{tabular}{l@{~}c@{~~}c@{~~}c@{~~}c@{~~}c@{~~}c@{~~}c@{~~}c@{~~}c@{~~}c@{~~}}
    
	\toprule
	
	& KCF & GOTURN  & ECO & RE3 & SiamFC  & MDNet & SiamRPN++ &\textbf{A3CT} & \textbf{A3CTD} \\
	& \cite{Henriques2014} & \cite{Held2016} & \cite{Danelljan2017} & \cite{Gordon2018} & \cite{Bertinetto2016} & \cite{Nam2015} & \cite{Li2018b} & &\\
	
	\midrule

	SS \quad & 0.178 & 0.214  & 0.324 & 0.325 & 0.336 & 0.397 & 0.496 & \textbf{0.306} & \textbf{0.415}\\
	PS \quad & 0.166 & 0.175  & 0.301 & 0.301 & 0.339 & 0.373 & - &\textbf{0.246} & \textbf{0.368}\\
	
	\bottomrule
    
    \end{tabular}
	}\vspace{1mm}%
\scriptsize

    \caption{State-of-the-art comparison on the LaSOT benchmark in terms of success score (SS) and precision score (PS). }
    \label{tab:lasotsota}%
	\vspace{-1mm}
\end{table}


\subsection{UAV123}
The UAV123 \cite{UAV123} is a benchmark composed of 123 videos acquired from low-altitude UAVs. The dataset is inherently different from traditional visual tracking benchmarks like OTB and VOT, since it offers sequences with an aerial point of view. 
 To evaluate our trackers, we use the same methodology and metrics used for the OTB-100 experiments.  

In Table \ref{tab:uav123sota} we present the scores against state-of-the-art trackers. A3CT performs 14\%, 8.2\% and 5.6\% better, in terms of SS, than KCF \cite{Henriques2014}, GOTURN \cite{Held2016} and ACT \cite{Chen2018} respectively. A3CTD has a 9.4\% SS and a 13.2\% PS improvements than A3CT and these lead to outperform SiamFC \cite{Bertinetto2016}, ECO \cite{Danelljan2017} and MDNet \cite{Nam2015} with a gain of, respectively, 4.2\%, 4\%, 3.7\% in SS and 2.4\%, 1.3\%, 0.7\% in PS.

\begin{table}[!h]
	\centering\vspace{-1mm}
	\resizebox{1.01\columnwidth}{!}{%
		\begin{tabular}{l@{~}c@{~~}c@{~~}c@{~~}c@{~~}c@{~~}c@{~~}c@{~~}c@{~~}c@{~~}c@{~~}}
    
	\toprule
	
	& KCF & GOTURN & ACT & RE3 & SiamFC & ECO & MDNet & SiamRPN++ &\textbf{A3CT} & \textbf{A3CTD} \\
	& \cite{Henriques2014} & \cite{Held2016} & \cite{Chen2018} & \cite{Gordon2018} & \cite{Bertinetto2016} & \cite{Danelljan2017} & \cite{Nam2015} & \cite{Li2018b} & &\\
	
	\midrule

	SS \quad & 0.331 & 0.389 & 0.415 & 0.514 & 0.523 & 0.525 & 0.528 & 0.613 & \textbf{0.471} & \textbf{0.565} \\
	PS \quad & 0.523 & 0.548 & 0.636 & 0.667 & 0.730 & 0.741 & 0.747 & 0.807 &\textbf{0.622} & \textbf{0.754} \\
	
	\bottomrule
	
    \end{tabular}
	}\vspace{1mm}%
\scriptsize

    \caption{State-of-the-art comparison on the UAV123 benchmark in terms of success score (SS) and precision score (PS).}
    \label{tab:uav123sota}%
	\vspace{-1mm}
\end{table}
\begin{figure}[!h]
\begin{center}
\includegraphics[width=.75\linewidth]{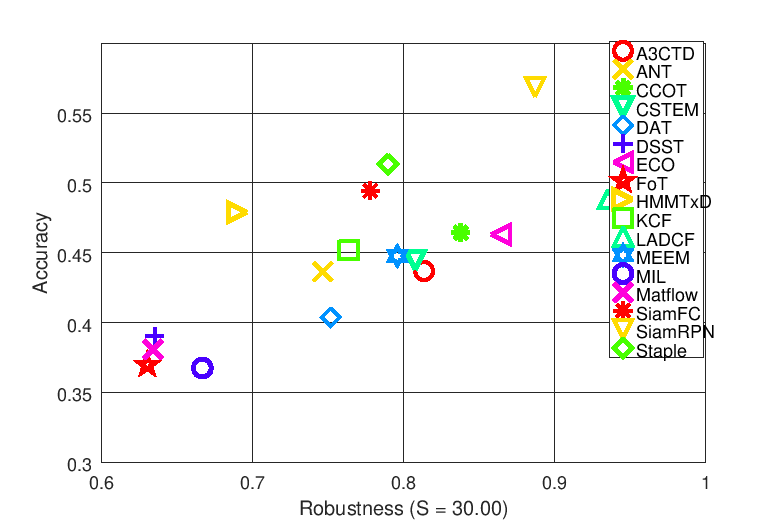}
\end{center}
   \caption{Accuracy-Robustness plot against some of the VOT-2018 \cite{VOT2018} competitors.}
\label{fig:vot2018ar}
\end{figure}
\begin{figure*}[!h]
\begin{center}
\includegraphics[width=.85\linewidth]{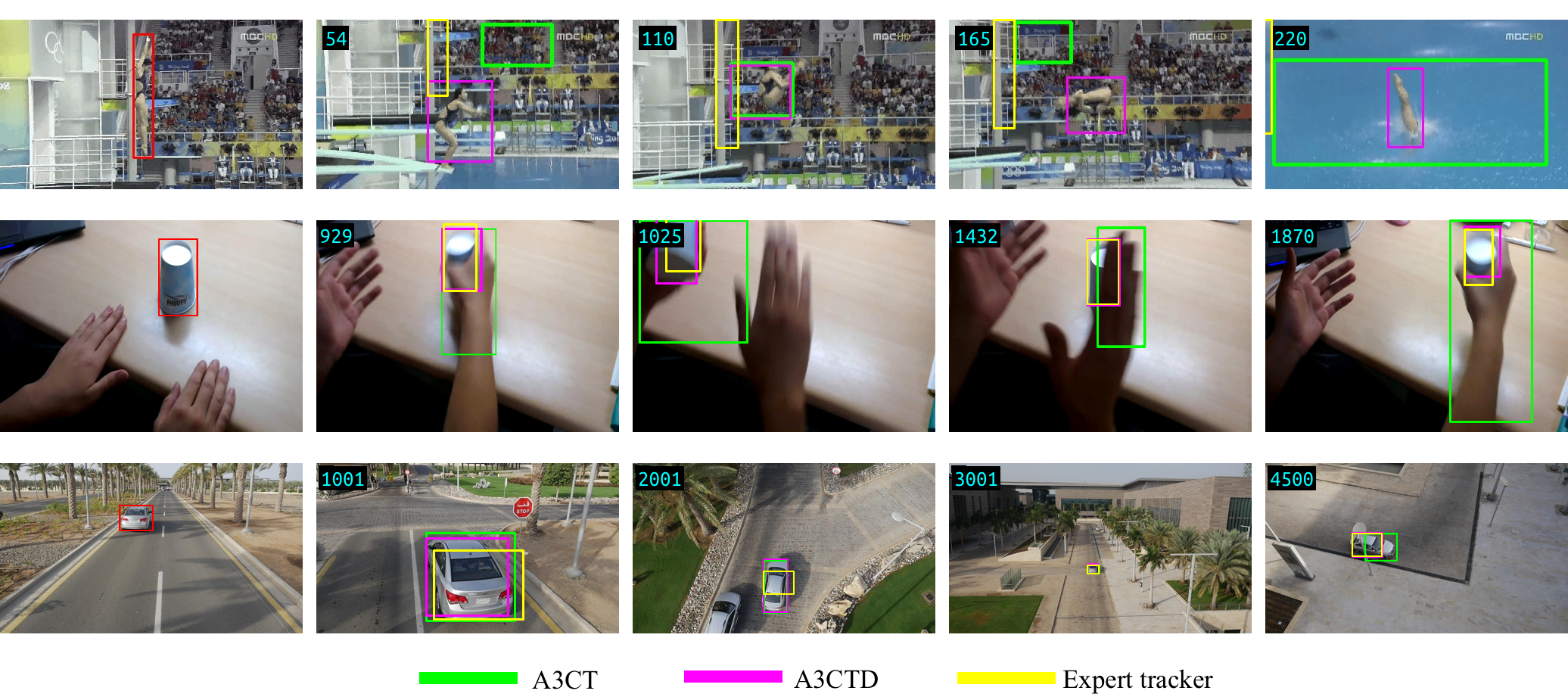}
\end{center}
   \caption{Qualitative examples of A3CT and A3CTD performance.}
\label{fig:qualitativeex}
\end{figure*}


\subsection{VOT benchmarks}
The VOT benchmarks are datasets used in the annual VOT tracking competition. These sets change year by year, introducing challenging tracking scenarios and increasing the difficulty of the task. 
Within the framework used by the VOT committee, trackers are evaluated based on Expected Average Overlap (EAO), Accuracy (A) and Robustness (R) ~\cite{VOT}. 
 We performed experiments on the test sets of VOT-2018 and VOT-2019 challenges. Both two benchmarks provide 60 (non completely overlapping) challenging videos.

\paragraph{VOT-2018.} In Figure \ref{fig:vot2018ar} we present the Accuracy-Robustness plot including A3CTD's performance in comparison with some of the partecipants to the VOT-2018 challenge. A3CTD achieves an EAO of 0.1847, an accuracy of 0.4536 while it failed (i.e. the IoU with the ground-truth becomes zero) 34.89 times. Our method perform definitely worse than the best solutions LADCF \cite{Xu2019}, SiamRPN \cite{Li2018} and ECO \cite{Danelljan2017} that achieved an EAO of 0.3889, 0.3837 and 0.2809 respectively. A3CTD's performance is however comparable to the one of SiamFC \cite{Bertinetto2016}, which achieved an EAO of 0.1875.

\paragraph{VOT-2019.} At the time of writing, the results of VOT-2019 challenge are not available. 
We submitted just the A3CTD tracker, since it resulted in the best performance generally. It achieved an EAO of 0.1652 and of 0.1497 for the \emph{baseline} and \emph{realtime} experiments respectively. The overlap in the \emph{baseline} experiment resulted in 0.4510.

\subsection{Ablation Study}
\begin{figure}[!h]
\begin{center}
\includegraphics[width=.85\linewidth]{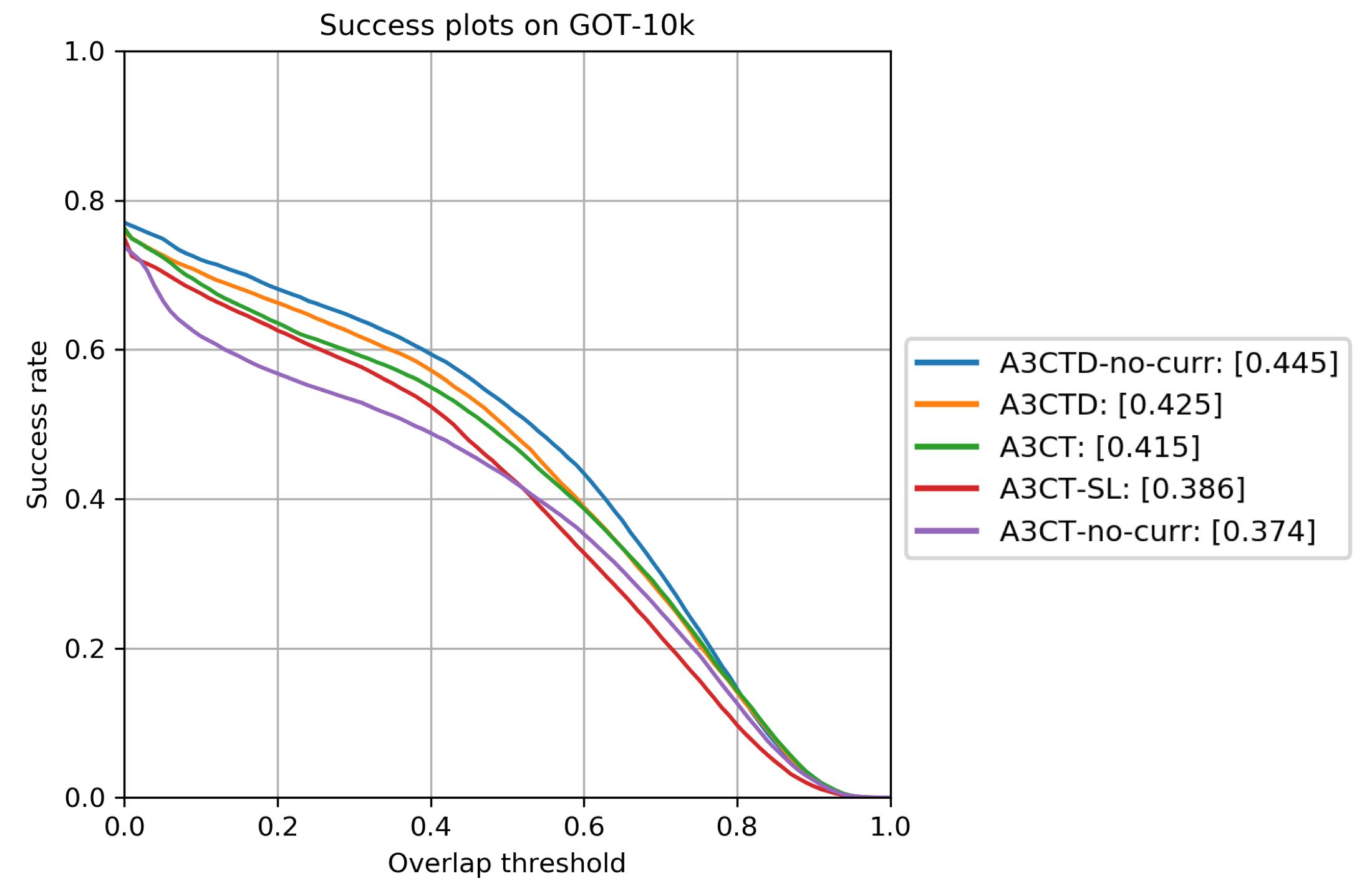}
\end{center}
   \caption{Success plot for the ablation study of A3CT and A3CTD.}
\label{fig:ablation}
\end{figure}

To assess the validity of all the features of our proposed solution we performed an ablation study on the GOT-10k test set. In particular, we ran experiments where we trained A3CT and A3CTD without the curriculum strategy (A3CT-no-curr and A3CTD-no-curr respectively) and A3CT with just imitating agents (A3CT-SL). In Figure \ref{fig:ablation} we report the success plot with the comparison of the different models involved. A3CT-SL performs worse than A3CT, suggesting that the use of RL agents is crucial to improve the baseline behaviour learned by the imitating agents. Moreover, since the state value function is learned by RL agents, this setup does not allow to exploit the demonstrator in the tracking phase.
A3CT-no-curr performs comparably to A3CT-SL, 4.1\% lower than A3CT. The curriculum learning strategy allows the tracking agent to learn a more precise tracking policy. Interestingly, A3CTD-no-curr outperforms A3CTD by 2\%. We believe that the increased length of the sequences during training allows the learning of a more accurate state value function, which is then able to make better predictions about the future behaviours of A3CT and the expert tracker. However, we chose A3CT and A3CTD as our final solution because of their lower difference in performance.

In terms of processing speed, A3CT runs at 90 FPS while A3CTD runs at 50 FPS. 


Finally, in Figure \ref{fig:qualitativeex} we present some qualitative examples of the tracking performance of A3CT and A3CTD.

\section{Conclusions and Future Work}
Thanks to the availability of a great amount of visual tracker and inspired by recent trends in RL, in this paper we proposed two novel trackers that are built on a deep regression network \cite{Held2016,Gordon2018}. The state-of-the-art tracking algorithm SiamFC \cite{Bertinetto2016} was executed on the large-scale tracking dataset GOT-10k \cite{GOT10k} to obtain expert demonstrations. The proposed network was then trained inside an RL on-policy asynchronous Actor-Critic framework \cite{Mnih2016} that incorporated parallel SL agents. Experiments showed that the proposed A3CT and A3CTD trackers outperform state-of-the-art methods on the most recent the GOT-10k test set \cite{GOT10k}, LaSOT \cite{LaSOT}, UAV123 \cite{UAV123} benchmarks, and perform comparably with the state-of-the-art on OTB-2015 \cite{OTB} and VOT benchmarks \cite{VOT2018}. Moreover, A3CT and A3CTD achieved a processing speed of 90 and 50 FPS respectively and thus are suitable for real-time applications.

Future works will focus on the integration of more expert trackers. In particular, we will study how the performance of our proposed trackers change when different expert trackers and when pools of experts are considered as demonstrators.


{\small
\bibliographystyle{ieee}
\bibliography{egbib}

\begin{thebibliography}{10}\itemsep=-1pt

\bibitem{Bengio2009}
Y.~Bengio, J.~Louradour, R.~Collobert, and J.~Weston.
\newblock {Curriculum learning}.
\newblock In {\em Proceedings of the 26th Annual International Conference on
  Machine Learning - ICML '09}, pages 1--8, New York, New York, USA, 2009. ACM
  Press.

\bibitem{Bertinetto2016}
L.~Bertinetto, J.~Valmadre, J.~F. Henriques, A.~Vedaldi, and P.~H.~S. Torr.
\newblock {Fully-Convolutional Siamese Networks for Object Tracking}.
\newblock jun 2016.

\bibitem{Chen2018}
B.~Chen, D.~Wang, P.~Li, S.~Wang, and H.~Lu.
\newblock {Real-time 'Actor-Critic' Tracking}.
\newblock In {\em The European Conference on Computer Vision (ECCV)}, 2018.

\bibitem{Choi2017}
J.~Choi, J.~Kwon, and K.~M. Lee.
\newblock {Visual Tracking by Reinforced Decision Making}.
\newblock {\em CoRR}, abs/1702.0, 2017.

\bibitem{Danelljan2017}
M.~Danelljan, G.~Bhat, F.~S. Khan, and M.~Felsberg.
\newblock {ECO: Efficient Convolution Operators for Tracking}.
\newblock In {\em International Conference on Computer Vision and Pattern
  Recgnition}, nov 2017.

\bibitem{Danelljan2019}
M.~Danelljan, G.~Bhat, F.~S. Khan, and M.~Felsberg.
\newblock {ATOM: Accurate Tracking by Overlap Maximization}.
\newblock In {\em International Conference on Computer Vision and Pattern
  Recognition}, nov 2018.

\bibitem{Danelljan2016}
M.~Danelljan, A.~Robinson, F.~S. Khan, and M.~Felsberg.
\newblock {Beyond Correlation Filters: Learning Continuous Convolution
  Operators for Visual Tracking}.
\newblock In {\em European Conference on Computer Vision}, aug 2016.

\bibitem{ImageNet}
J.~Deng, W.~Dong, R.~Socher, L.-J. Li, {Kai Li}, and {Li Fei-Fei}.
\newblock {ImageNet: A large-scale hierarchical image database}.
\newblock In {\em 2009 IEEE Conference on Computer Vision and Pattern
  Recognition}, pages 248--255. IEEE, jun 2009.

\bibitem{LaSOT}
H.~Fan, L.~Lin, F.~Yang, P.~Chu, G.~Deng, S.~Yu, H.~Bai, Y.~Xu, C.~Liao, and
  H.~Ling.
\newblock {LaSOT: A High-quality Benchmark for Large-scale Single Object
  Tracking}.
\newblock In {\em International Conference on Computer Vision and Pattern
  Recognition}, sep 2019.

\bibitem{Gordon2018}
D.~Gordon, A.~Farhadi, and D.~Fox.
\newblock {Re3: Real-Time Recurrent Regression Networks for Visual Tracking of
  Generic Objects}.
\newblock {\em IEEE Robotics and Automation Letters}, 3(2):788--795, 2018.

\bibitem{Guo2017}
Q.~Guo, W.~Feng, C.~Zhou, R.~Huang, L.~Wan, and S.~Wang.
\newblock {Learning Dynamic Siamese Network for Visual Object Tracking}.
\newblock In {\em 2017 IEEE International Conference on Computer Vision
  (ICCV)}, pages 1781--1789. IEEE, oct 2017.

\bibitem{He2016}
K.~He, X.~Zhang, S.~Ren, and J.~Sun.
\newblock {Deep residual learning for image recognition}.
\newblock In {\em International Conference on Computer Vision and Pattern
  Recognition}, pages 770--778, 2016.

\bibitem{Held2016}
D.~Held, S.~Thrun, and S.~Savarese.
\newblock {Learning to Track at 100 {\{}FPS{\}} with Deep Regression Networks}.
\newblock {\em European Conference on Computer Vision}, abs/1604.0, 2016.

\bibitem{Henriques2014}
J.~F. Henriques, R.~Caseiro, P.~Martins, and J.~Batista.
\newblock {High-Speed Tracking with Kernelized Correlation Filters}.
\newblock {\em CoRR}, abs/1404.7, 2014.

\bibitem{Hester2018}
T.~Hester, M.~Vecerik, O.~Pietquin, M.~Lanctot, T.~Schaul, B.~Piot, D.~Horgan,
  J.~Quan, A.~Sendonaris, G.~Dulac-Arnold, I.~Osband, J.~Agapiou, J.~Z. Leibo,
  and A.~Gruslys.
\newblock {Deep Q-learning from Demonstrations}.
\newblock In {\em AAAI}, apr 2018.

\bibitem{Hochreiter1997}
S.~Hochreiter and J.~Schmidhuber.
\newblock {Long Short-Term Memory}.
\newblock {\em Neural Comput.}, 9(8):1735--1780, nov 1997.

\bibitem{Huang2017}
C.~Huang, S.~Lucey, and D.~Ramanan.
\newblock {Learning Policies for Adaptive Tracking with Deep Feature Cascades}.
\newblock In {\em 2017 IEEE International Conference on Computer Vision
  (ICCV)}, pages 105--114. IEEE, oct 2017.

\bibitem{GOT10k}
L.~Huang, X.~Zhao, and K.~Huang.
\newblock {GOT-10k: A Large High-Diversity Benchmark for Generic Object
  Tracking in the Wild}.
\newblock oct 2018.

\bibitem{Supancic2017}
J.~S.~S. III and D.~Ramanan.
\newblock {Tracking as Online Decision-Making: Learning a Policy from Streaming
  Videos with Reinforcement Learning}.
\newblock {\em CoRR}, abs/1707.0, 2017.

\bibitem{Kang2018}
B.~Kang, Z.~Jie, and J.~Feng.
\newblock {Policy Optimization with Demonstrations}.
\newblock In {\em ICML 2018}, volume~80, pages 2474--2483, 2018.

\bibitem{Kingma2014}
D.~P. Kingma and J.~Ba.
\newblock {Adam: {\{}A{\}} Method for Stochastic Optimization}.
\newblock {\em CoRR}, abs/1412.6, 2014.

\bibitem{Konda2000}
V.~R. Konda and J.~N. Tsitsiklis.
\newblock {Actor-Critic Algorithms}.
\newblock In {\em Advances in Neural Information Processing Systems}, 2000.

\bibitem{VOT2015}
J.~. L. A. . F. M. . C. L. . F. G. . V. T. . H. G. . N. G. .~P. Kristan, Matej
  ;~Matas.

\bibitem{VOT2016}
M.~Kristan, A.~Leonardis, J.~Matas, M.~Felsberg, R.~Pflugfelder,
  L.~{\v{C}}ehovin, T.~Voj{\'{i}}r̃, G.~H{\"{a}}ger, A.~Luke{\v{z}}i{\v{c}},
  G.~Fern{\'{a}}ndez, et~al.
\newblock {The Visual Object Tracking VOT2016 Challenge Results}.
\newblock In {\em European Conference on Computer Vision}, pages 777--823.
  Springer, Cham, 2016.

\bibitem{VOT2017}
M.~Kristan, A.~Leonardis, J.~Matas, M.~Felsberg, R.~Pflugfelder, L.~C. Zajc,
  T.~Vojir, et~al.
\newblock {The Visual Object Tracking VOT2017 Challenge Results}.
\newblock In {\em 2017 IEEE International Conference on Computer Vision
  Workshops (ICCVW)}, pages 1949--1972. IEEE, oct 2017.

\bibitem{VOT2018}
M.~Kristan, A.~Leonardis, J.~Matas, M.~Felsberg, R.~Pflugfelder, L.~{\v{C}}.
  Zajc, T.~Voj{\'{i}}r̃, G.~Bhat, A.~Luke{\v{z}}i{\v{c}}, A.~Eldesokey,
  G.~Fern{\'{a}}ndez, et~al.
\newblock {The Sixth Visual Object Tracking VOT2018 Challenge Results}.
\newblock In {\em European Conference on Computer Vision}, pages 3--53.
  Springer, Cham, sep 2019.

\bibitem{VOT}
M.~Kristan, J.~Matas, A.~Leonardis, T.~Vojir, R.~Pflugfelder, G.~Fernandez,
  G.~Nebehay, F.~Porikli, and L.~{\v{C}}ehovin.
\newblock {A Novel Performance Evaluation Methodology for Single-Target
  Trackers}.
\newblock {\em IEEE Transactions on Pattern Analysis and Machine Intelligence},
  38(11):2137--2155, nov 2016.

\bibitem{VOT2014}
M.~Kristan, R.~Pflugfelder, A.~Leonardis, J.~Matas, L.~{\v{C}}ehovin,
  G.~Nebehay, T.~Voj{\'{i}}ř, G.~Fern{\'{a}}ndez, A.~Luke{\v{z}}i{\v{c}},
  et~al.
\newblock {The Visual Object Tracking VOT2014 Challenge Results}.
\newblock In {\em European Conference on Computer Vision}, pages 191--217.
  Springer, Cham, 2015.

\bibitem{Lecun98}
Y.~Lecun, L.~Bottou, Y.~Bengio, and P.~Haffner.
\newblock {Gradient-based learning applied to document recognition}.
\newblock In {\em Proceedings of the IEEE}, pages 2278--2324, 1998.

\bibitem{Li2018b}
B.~Li, W.~Wu, Q.~Wang, F.~Zhang, J.~Xing, and J.~Yan.
\newblock {SiamRPN++: Evolution of Siamese Visual Tracking with Very Deep
  Networks}.
\newblock dec 2018.

\bibitem{Li2018}
B.~Li, J.~Yan, W.~Wu, Z.~Zhu, and X.~Hu.
\newblock {High Performance Visual Tracking with Siamese Region Proposal
  Network}.
\newblock In {\em 2018 IEEE/CVF Conference on Computer Vision and Pattern
  Recognition}, pages 8971--8980. IEEE, jun 2018.

\bibitem{Mnih2016}
V.~Mnih, A.~P. Badia, M.~Mirza, A.~Graves, T.~P. Lillicrap, T.~Harley,
  D.~Silver, and K.~Kavukcuoglu.
\newblock {Asynchronous Methods for Deep Reinforcement Learning}.
\newblock {\em CoRR}, abs/1602.0, 2016.

\bibitem{Mnih2013}
V.~Mnih, K.~Kavukcuoglu, D.~Silver, A.~Graves, I.~Antonoglou, D.~Wierstra, and
  M.~A. Riedmiller.
\newblock {Playing Atari with Deep Reinforcement Learning}.
\newblock {\em CoRR}, abs/1312.5, 2013.

\bibitem{Mnih2015}
V.~Mnih, K.~Kavukcuoglu, D.~Silver, A.~A. Rusu, J.~Veness, M.~G. Bellemare,
  A.~Graves, M.~Riedmiller, A.~K. Fidjeland, G.~Ostrovski, S.~Petersen,
  C.~Beattie, A.~Sadik, I.~Antonoglou, H.~King, D.~Kumaran, D.~Wierstra,
  S.~Legg, and D.~Hassabis.
\newblock {Human-level control through deep reinforcement learning}.
\newblock {\em Nature}, 518(7540):529--533, feb 2015.

\bibitem{UAV123}
M.~Mueller, N.~Smith, and B.~Ghanem.
\newblock {A Benchmark and Simulator for UAV Tracking}.
\newblock In {\em European Conference on Computer Vision}, pages 445--461.
  Springer, Cham, 2016.

\bibitem{Nair2018}
A.~Nair, B.~McGrew, M.~Andrychowicz, W.~Zaremba, and P.~Abbeel.
\newblock {Overcoming Exploration in Reinforcement Learning with
  Demonstrations}.
\newblock In {\em Proceedings - IEEE International Conference on Robotics and
  Automation}, pages 6292--6299. Institute of Electrical and Electronics
  Engineers Inc., sep 2018.

\bibitem{Nam2015}
H.~Nam and B.~Han.
\newblock {Learning Multi-Domain Convolutional Neural Networks for Visual
  Tracking}.
\newblock {\em CoRR}, abs/1510.0, 2015.

\bibitem{Paszke2017}
A.~Paszke, S.~Gross, S.~Chintala, G.~Chanan, E.~Yang, Z.~DeVito, Z.~Lin,
  A.~Desmaison, L.~Antiga, and A.~Lerer.
\newblock {Automatic differentiation in PyTorch}.
\newblock 2017.

\bibitem{Ren2018}
L.~Ren, X.~Yuan, J.~Lu, M.~Yang, and J.~Zhou.
\newblock {Deep Reinforcement Learning with Iterative Shift for Visual
  Tracking}.
\newblock In {\em The European Conference on Computer Vision (ECCV)}, 2018.

\bibitem{Salimans2018}
T.~Salimans and R.~Chen.
\newblock {Learning Montezuma's Revenge from a Single Demonstration}.
\newblock In {\em Proceedings of NeurIPS 2018 Workshop on Deep RL}, dec 2018.

\bibitem{Silver2016}
D.~Silver, A.~Huang, C.~J. Maddison, A.~Guez, L.~Sifre, G.~van~den Driessche,
  J.~Schrittwieser, I.~Antonoglou, V.~Panneershelvam, M.~Lanctot, S.~Dieleman,
  D.~Grewe, J.~Nham, N.~Kalchbrenner, I.~Sutskever, T.~Lillicrap, M.~Leach,
  K.~Kavukcuoglu, T.~Graepel, and D.~Hassabis.
\newblock {Mastering the Game of {\{}Go{\}} with Deep Neural Networks and Tree
  Search}.
\newblock {\em Nature}, 529(7587):484--489, 2016.

\bibitem{Silver2017}
D.~Silver, J.~Schrittwieser, K.~Simonyan, I.~Antonoglou, A.~Huang, A.~Guez,
  T.~Hubert, L.~Baker, M.~Lai, A.~Bolton, Y.~Chen, T.~Lillicrap, F.~Hui,
  L.~Sifre, G.~van~den Driessche, T.~Graepel, and D.~Hassabis.
\newblock {Mastering the game of Go without human knowledge}.
\newblock {\em Nature}, 550:354----, 2017.

\bibitem{Smeulders2014}
A.~W.~M. Smeulders, D.~M. Chu, R.~Cucchiara, S.~Calderara, A.~Dehghan, and
  M.~Shah.
\newblock {Visual Tracking: An Experimental Survey}.
\newblock {\em IEEE Transactions on Pattern Analysis and Machine Intelligence},
  36(7):1442--1468, 2014.

\bibitem{Song2017}
Y.~Song, C.~Ma, L.~Gong, J.~Zhang, R.~Lau, and M.-H. Yang.
\newblock {CREST: Convolutional Residual Learning for Visual Tracking}.
\newblock In {\em Internationa{\`{o}} Conference on Computer Vision}, aug 2017.

\bibitem{SuttonBarto2018}
R.~S. Sutton and A.~G. Barto.
\newblock {\em {Reinforcement Learning: An Introduction}}.
\newblock MIT Press, Cambridge, MA, USA, 2nd edition, 2018.

\bibitem{Valmadre2017}
J.~Valmadre, L.~Bertinetto, J.~Henriques, A.~Vedaldi, and P.~H.~S. Torr.
\newblock {End-to-End Representation Learning for Correlation Filter Based
  Tracking}.
\newblock In {\em 2017 IEEE Conference on Computer Vision and Pattern
  Recognition (CVPR)}, pages 5000--5008. IEEE, jul 2017.

\bibitem{Vecerik2017}
M.~Vecerik, T.~Hester, J.~Scholz, F.~Wang, O.~Pietquin, B.~Piot, N.~Heess,
  T.~Roth{\"{o}}rl, T.~Lampe, and M.~Riedmiller.
\newblock {Leveraging Demonstrations for Deep Reinforcement Learning on
  Robotics Problems with Sparse Rewards}.
\newblock jul 2017.

\bibitem{Watkins1992}
C.~J. C.~H. Watkins and P.~Dayan.
\newblock {Q-learning}.
\newblock {\em Machine Learning}, 8(3):279--292, 1992.

\bibitem{Williams1992}
R.~J. Williams.
\newblock {Simple Statistical Gradient-Following Algorithms for Connectionist
  Reinforcement Learning}.
\newblock {\em Mach. Learn.}, 8(3-4):229--256, may 1992.

\bibitem{OTB}
Y.~Wu, J.~Lim, and M.-H. Yang.
\newblock {Online Object Tracking: A Benchmark.}
\newblock In {\em CVPR}, pages 2411--2418. IEEE Computer Society, 2013.

\bibitem{Xu2019}
T.~Xu, Z.-H. Feng, X.-J. Wu, and J.~Kittler.
\newblock {Learning Adaptive Discriminative Correlation Filters via Temporal
  Consistency Preserving Spatial Feature Selection for Robust Visual Tracking}.
\newblock {\em IEEE Transactions on Image Processing}, jul 2019.

\bibitem{Yun2017}
S.~Yun, J.~Choi, Y.~Yoo, K.~Yun, and J.~Y. Choi.
\newblock {Action-Decision Networks for Visual Tracking with Deep Reinforcement
  Learning}.
\newblock In {\em 2017 IEEE Conference on Computer Vision and Pattern
  Recognition (CVPR)}, pages 1349--1358. IEEE, jul 2017.

\bibitem{Zhang2019}
Z.~Zhang and H.~Peng.
\newblock {Deeper and Wider Siamese Networks for Real-Time Visual Tracking}.
\newblock In {\em International Conference on Computer Vision and Pattern
  Recognition}, jan 2019.

\bibitem{Zhu2018}
Z.~Zhu, Q.~Wang, B.~Li, W.~Wu, J.~Yan, and W.~Hu.
\newblock {Distractor-aware Siamese Networks for Visual Object Tracking}.
\newblock In {\em European Conference on Computer Vision}, aug 2018.

\end{thebibliography}
}

\end{document}